\documentclass[final]{cvpr}

\usepackage{times}
\usepackage{epsfig}
\usepackage{graphicx}
\usepackage{amsmath}
\usepackage{amssymb}
\usepackage{tabularx,booktabs}

\usepackage[pagebackref=true,breaklinks=true,colorlinks,bookmarks=false]{hyperref}

\def\cvprPaperID{5457} 
\def\confYear{CVPR 2021}

\begin{document}


\title{Continuous Face Aging via Self-estimated Residual Age Embedding}

\author{Zeqi Li\thanks{This work is done during Zeqi Li’s full-time employment at ModiFace.}\\
ModiFace\\
{\tt\small lizeqi@cs.toronto.edu}
\and
Ruowei Jiang\\
ModiFace\\
{\tt\small irene@modiface.com}
\and
Parham Aarabi\\
ModiFace\\
{\tt\small parham@modiface.com}
}

\maketitle

\begin{abstract}

Face synthesis, including face aging, in particular, has been one of the major topics that witnessed a substantial improvement in image fidelity by using generative adversarial networks (GANs). Most existing face aging approaches divide the dataset into several age groups and leverage group-based training strategies, which lacks the ability to provide fine-controlled continuous aging synthesis in nature. In this work, we propose a unified network structure that embeds a linear age estimator into a GAN-based model, where the embedded age estimator is trained jointly with the encoder and decoder to estimate the age of a face image and provide a personalized target age embedding for age progression/regression. The personalized target age embedding is synthesized by incorporating both personalized residual age embedding of the current age and exemplar-face aging basis of the target age, where all preceding aging bases are derived from the learned weights of the linear age estimator. This formulation brings the unified perspective of estimating the age and generating personalized aged face, where self-estimated age embeddings can be learned for every single age. The qualitative and quantitative evaluations on different datasets further demonstrate the significant improvement in the continuous face aging aspect over the state-of-the-art.
\end{abstract}


\section{Introduction}
Face aging, also known as age progression, aims to aesthetically render input face images with natural aging and rejuvenating effects while preserving identity information of the individual. With recent advances in deep learning, face synthesis has also shown substantial improvement on image fidelity and the age precision in the simulated face images \cite{he2019s2gan,wang2018face,orel2020lifespan}. A major challenge to solve a variety of remaining problems (e.g. continuous aging) is the lack of data. For example, many research works of face aging \cite{liu2017face,wang2018face,yang2018learning,he2019s2gan} need to group images into 4-5 age groups (such as \(<\)30, 30-40, 40-50, 50+) and can only generate images within a target age group, due to the limited amount of data at each age. Another important problem is how to maintain personal traits in age progression, as aging patterns may differ for each individual. 

Traditional face aging contains mainly two approaches: physical model-based \cite{boissieux2000simulation,wu1994plastic} and prototype-based \cite{tiddeman2001prototyping,kemelmacher2014illumination}. The physical model-based methods often consist of complex physical modeling, considering skin wrinkles, face shape, muscle changes, and hair color, etc. This type of method typically requires a tremendous amount of data and is very expensive computationally. 
Prototype-based methods firstly explore group-based designs by computing an average face within the pre-defined age groups, which fails to retain personalized aging information. Further, all those methods are not applicable to continuous face aging.

Following the success of recent generative models, such as variational autoencoders (VAEs) and generative adversarial networks (GANs) \cite{goodfellow2014generative}, on the image translation tasks, researchers have dedicated efforts in adapting those methods to face synthesis. IPCGAN \cite{wang2018face} has shown significant progress in generating face images with evident aging effects by enforcing an age estimation loss.  Later variation \cite{yang2018learning} creates a pyramid structure for the discriminator to improve face aging understanding at multiple scales. Continuous aging was not explored among these methods. He et al. \cite{he2019s2gan} introduced a multi-branch generator for the group-based training and proposed the idea to approximate continuous aging via linear interpolation of latent representations between two adjacent age groups. The authors of \cite{orel2020lifespan} also tackle the problem using a similar linear interpolation approach, which is performed on the learned age latent code between two neighboring groups instead. These types of methods make an assumption that the age progression is linear between the two adjacent groups and the learned group embedding can be used directly as the median age embedding. Consequently, this may result in a shift of target age in the generated images. Intuitively, this nonlinearity can be interpreted as: people do not age at the same speed for different stages. Moreover, such interpolation-based methods may alter personal traits when disentanglement is imperfect. 

To address the aforementioned problems, we propose a novel approach to achieve continuous aging by a unified network where a simple age estimator is embedded into a regular encoder-decoder architecture.
This allows the network to learn self-estimated age embeddings of all ages, thus representing the continuous aging information without manual efforts in selecting proper anchor age groups.
Given a target age, we derive a personalized age embedding which considers two aspects of face aging: 1) a personalized residual age embedding at the current age, which preserves the individual's aging information; 2) exemplar-face aging basis at the target age, which encodes the shared aging patterns among the entire population. We describe the detailed calculation and training mechanism in \textbf{Method}. The calculated target age embedding is then used for final image generation. We experiment extensively on FFHQ \cite{ karras2019style} and CACD2000 \cite{chen2014cross} datasets. Our results, both qualitatively and quantitatively, show significant improvement over the state-of-the-art in various aspects. 
Our main contributions are:
\begin{itemize}
\item[$\bullet$] We propose a novel method to self-estimate continuous age embeddings and derive personalized age embeddings for face aging task by jointly training an age estimator with the generator. We quantitatively and qualitatively demonstrate that the generated images better preserve the personalized information, achieve more accurate aging control, and present more fine-grained aging details.
\item[$\bullet$] We show that our continuous aging approach generates images with more well-aligned target ages, and better preserves detailed personal traits, without manual efforts to define proper age groups.
\item[$\bullet$] Our proposed idea to self-estimate personalized age embedding from a related discriminative model can be easily applied to other conditional image-to-image translation tasks, without introducing extra complexity. In particular, tasks involving a continuous condition and modeling (e.g. non-smile to smile), can benefit from this setup.
\end{itemize}

\section{Related Work}
\subsection{Face Aging Model} 
Traditional methods can be categorized as physical model-based approaches \cite{boissieux2000simulation,wu1994plastic,suo2012concatenational} and prototype-based approaches \cite{rowland1995manipulating,tiddeman2001prototyping,kemelmacher2014illumination,lanitis2002toward}. The physical model-based methods focuses on creating models to address specific sub-effects of aging, such as skin wrinkles \cite{wu1994plastic,bando2002simple,boissieux2000simulation}, craniofacial growth \cite{todd1980perception,ramanathan2006modeling}, muscle structure \cite{suo2012concatenational,ramanathan2008modeling}, and face components \cite{suo2007multi,suo2009compositional}. These methods are often very complicated, which typically require a sequence of face images of the same person at different ages and expert knowledge of the aging mechanism. The prototype-based approaches \cite{rowland1995manipulating,tiddeman2001prototyping,burt1995perception} explore face progression problem using group-based learning where an average face is estimated within each age group. However, personalized aging patterns and identity information are not well-preserved in such strategies. In \cite{wang2016category,yang2016face,shu2015personalized}, sparse representation of the input image have been utilized to express personalized face transformation patterns. Though the personalized aging patterns are preserved to some extent by such approaches, the synthesized images suffer from quality issues.

Recently, deep learning approaches have been adopted to model personalized aging transformations. Wang et al. \cite{wang2016recurrent} proposed a recurrent neural network model, leveraging a series of recurrent forward passes for a more smooth transition from young to old. Later GAN-based works \cite{li2018global,wang2018face,yang2018learning} have shown superior breakthroughs on the fidelity of images. Li et al. \cite{li2018global} designed three subnets for local patches and fused local and global features to obtain a smooth synthesized image. IPCGAN \cite{wang2018face} enforces an age estimation loss on the generated image and an identity loss to achieve good face aging effects. More efforts have also been made to address age accuracy and identity permanence. Yang et al.\cite{yang2018learning} and Liu et al. \cite{liu2017face} introduce a modification of discriminator losses to guide a more accurate age of the output images. Authors of \cite{liu2019attribute} improved image quality of synthesized images by using a wavelet packet transformation and multiple facial attribute encoding. However, these methods \cite{wang2018face,yang2018learning,liu2017face} condition the output image by concatenating one-hot vector representing the target age groups. To obtain a continuous aging condition, the vector will be extended to a much larger dimension, which makes training unstable and more complicated. Furthermore, it requires a tremendous amount of training images.

Though some works \cite{zhang2017age,antipov2017face,shen2020interpreting}, which aim to interpolate features in the latent space, provided a direction to support continuous aging, they have limited ability to produce high-quality images while preserving the identity. In \cite{he2019s2gan}, the authors proposed to linear interpolate feature vectors from adjacent age groups upon group-based training to achieve continuous aging progression. Similarly, \cite{orel2020lifespan} linearly interpolates between two adjacent anchor age embeddings. These methods follow the assumption that the embeddings are aligned linearly between anchors, which makes the decision of anchor ages crucial. In this work, we present continuous self-estimated age embeddings free of manual efforts while achieving better continuous age modeling.

\subsection{Generative Adversarial Networks}
Generative adversarial networks \cite{goodfellow2014generative} have been a popular choice on image-to-image translations tasks. CycleGAN \cite{zhu2017unpaired} and Pix2Pix \cite{pix2pix2017} explored image translations between two domains using unpaired and paired training samples respectively. More recent works \cite{choi2018stargan,liu2019stgan} proposed training techniques to enable multi-domain translation. In \cite{mirza2014conditional, odena2017conditional}, authors firstly explored conditional image generation as extensions to basic GANs. Later works \cite{choi2020stargan,park2019gaugan} have further shown superiority on many conditional image translation tasks, by transforming and injecting the condition into the model in a more effective manner.

\subsection{Face Age Estimation}
The task to predict apparent age refers to the regression problem that estimates a continuous numerical value for each given face image. Deep Expectation of Apparent Age (DEX) \cite{rothe2015dex} proposed a method to achieve a MAE of 3.25 on MORPH II \cite{ricanek2006morph}, by combining classification loss and regression loss. Pan et al. \cite{pan2018mean} proposed to use mean-variance loss on the probability distribution to further improve the MAE to 2.16 on MORPH II.

\begin{figure}
\centering
\includegraphics[width=0.48 \textwidth]{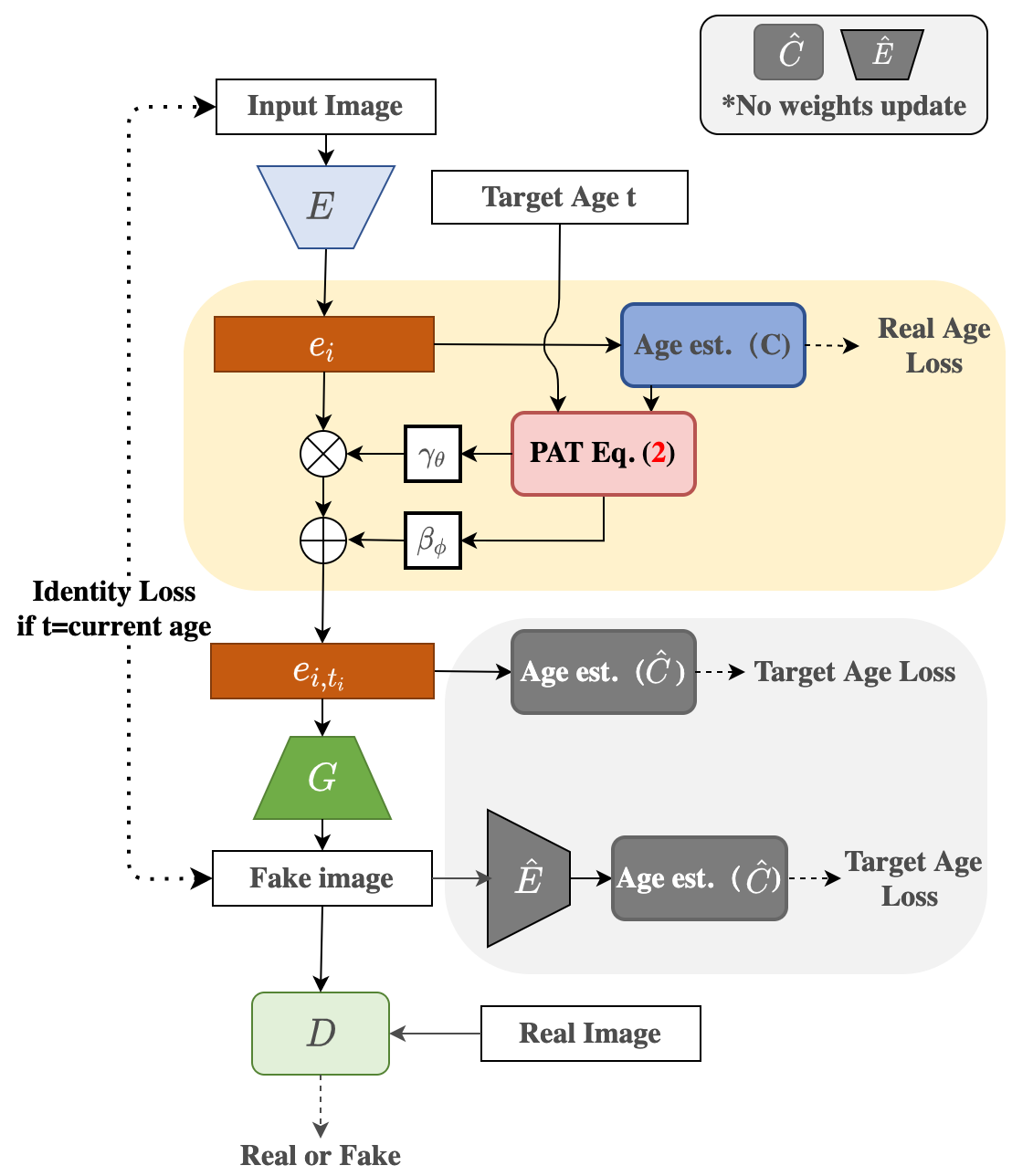}
\caption{Model architecture: An age estimator is jointly trained with an image generator, where \(\mathbf{E}\) is the shared encoder and \(\mathbf{C}\) is branched off for the age estimation task. The personalized age embedding transformation ($\mathbf{PAT}$, \textbf{Eq.} (\ref{personalized_age_basis})) is based on two components: 
1) residual aging basis at the current age; 
2) exemplar-face aging basis at the target age. Then the transformed identity encoding is decoded by $\mathbf{G}$. The whole model is learned with the age losses, identity loss, and the adversarial loss.}
\label{method_diag}
\end{figure}
\section{Method}
\subsection{Formulation}
As shown in \textbf{Fig.} \ref{method_diag}, our model consists of four components: 1) identity encoding module \(\mathbf{E}\); 2) age estimation module \(\mathbf{C}\); 3) personalized age embedding transformation module \(\mathbf{PAT}\); 4) aged face generation module \(\mathbf{G}\). During inference, we apply an encoder network \(\mathbf{E}\) to extract the identity information from the given image \(x_i\), where the encoding is denoted as $e_i = \mathbf{E}(x_i)$. Then an embedded age estimator \(\mathbf{C}\) is used to obtain the age probability distribution of the identity encoding. Based on the self-estimated age distribution and the target age \(t\), we apply a personalized age embedding transformation \(\mathbf{PAT}\) on the identity encoding \(e_i\). Lastly, the synthesized face is decoded from the transformed identity encoding \(\mathbf{PAT}(e_i,t)\) by the generator \(\mathbf{G}\). All modules are optimized jointly end-to-end under three objectives including the mean-variance age loss \cite{pan2018mean} for accurate aging, the \(L1\) reconstruction loss for identity preservation, and the adversarial loss for image realism. Unlike many prior face aging works \cite{wang2018face, he2019s2gan} in which require a pre-trained age classifier to guide the face aging training, our model directly obtains a self-estimated age embedding by utilizing a unified framework for achieving face aging and age estimation at the same time. More favorably, the embedded age estimator not only enables personalized continuous age transformation in a more accurate manner, compared to the interpolation-based approach, but also provides the guidance for face image generation.

\textbf{Identity Age Estimation Module (C)}
In prior works \cite{wang2018face, he2019s2gan}, face aging and face age estimation are treated as two independent tasks where an age estimation model, usually a classifier, is pre-trained separately and then used to guide the generator to realize natural aging effects. As the two mentioned tasks are intrinsically related, both goals can be achieved with one unified structure by sharing an encoder $\mathbf{E}$. The age estimator $\mathbf{C}$, in our case containing a global average pooling layer and a fully-connected layer, is branched off from $\mathbf{E}$. Finally, the age probability distribution \(p_i \in R^K\) can be obtained by performing the softmax function, where \(K\) denotes the number of age classes.
Without introducing too much extra complexity, such unified design also provides three advantages. Firstly, it eliminates the need to acquire a well-trained age estimator model beforehand. Secondly, age estimation on the identity encoding helps the model to establish a more age-specific identity representation. Thirdly, the weight \(W_C\) in the fully-connected layer is also used as the age embedding bases (bias terms are set to zero) which encodes the exemplar-face information from a metric learning perspective. In notation:

\begin{equation}
a_j= W_C[j],
\label{age_basis}
\end{equation}
where \(W_C \in \mathbb{R}^{K \times D}\),\(a_j \in \mathbb{R}^D\) and \(D\) equals to the channel dimension of the identity encoding.

\textbf{Personalized Age embedding Transformation (PAT)}
Face aging is a challenging and ambiguous task in nature as different facial signs/symptoms ages differently for different people at different stages. Thus, personalization is desired in performing face aging. In our design, we characterize this personalization by a residual age embedding calculated from the age probability distribution \(p_{i} \in \mathbb{R}^K\) and the exemplar-face aging basis \(a_{j} \in \mathbb{R}^D\) where \(i\) denotes the sample \(i\) and \(j \in {1,2,…,K}\) denotes the age. \(p_{i,j} \in R\) is the probability at age j for sample i. To obtain the personalized aging basis for any target age \(t_i\), we formulate the process as the following operation:

\begin{equation}
\tilde{a}_{i,t_i} = (\sum_{j=1}^{K} p_{i,j}a_{j} - a_{j=[m_i]}) + a_{j=t_i},
\label{personalized_age_basis}
\end{equation}

The \(\sum_{j=1}^{K} p_{i,j}a_{j}\) term represents the personalized aging basis of the identity by taking the expected value of the aging basis based on the age probability distribution. Then we can obtain the residual age embedding by subtracting the exemplar-face aging basis at the current (self-estimated) age \(a_{j=[m_i]}\) from the personalized aging basis. The residual age embedding preserves the identity’s personalized factors while removing the prevailing aging factors at the self-estimated age. The final personalized target age embedding \(\tilde{a}_{i,t_i}\) is obtained by adding the exemplar-face aging basis at the target aging basis \(a_{j=t_i}\), which encodes the shared aging factors at the target age among the entire population. 
With the personalized target age embedding \(\tilde{a}_{i,t_i}\), we then apply an affine projection transformation to derive the scale and shift coefficients for the original identity encoding \(E(x_i)=e_i\), similar to Conditional BN \cite{NIPS2017_7237} and AdaIN \cite{huang2017adain}:

\begin{equation}
\mathbf{PAT}(e_i, t_i) = e_{i, t_i} = \gamma_\theta(\tilde{a}_{i,t_i}) e_i + \beta_\phi(\tilde{a}_{i,t_i}),
\label{personalized_age_transformation}
\end{equation}

In our experiments, we do not observe significant performance difference w/wo \(\beta_\phi(\tilde{a}_{i,t_i})\).

\textbf{Continuous Aging}
As the aging bases from the fully-connected layer encode every single age, any integer target age is naturally supported. While some previous group-based approaches only model a few anchor age groups and achieving continuous aging via linear interpolation in the latent space. Our proposed method, however, explicitly models a fine-controlled age progression for each age and also supports float target age via a weighted sum of 2
neighboring integer age embedding bases.

\subsection{Objective}
The design of the objectives ensures the synthesized face image reflects accurate age progression/regression, preserves the identity, and looks realistic.

\textbf{Mean-Variance Age Loss}
The age loss plays two roles in our network: 1) it helps the estimator learn good aging bases for all ages; 2) it guides the generator by estimating the age of the generated fake images. To achieve both goals, we adopt the mean-variance age loss proposed by \cite{pan2018mean}. Given an input image \(x_i\) and an age label \(y_i\), the mean-variance loss is defined as below:

\begin{equation}
\begin{split}
\mathbf{L}_{mv} &= L_s + \lambda_{mv1} L_m + \lambda_{mv2} L_v \\ 
                                &= \frac{1}{N} \sum_{i=1}^{N} -log p_{i, y_i} + \frac{\lambda_1}{2} (m_i - y_i)^2 + \lambda_2 v_i,
\label{mean_variance_loss}
\end{split}
\end{equation}

where \(m_i = \sum_{j=1}^{K} jp_{i,j}\) is the mean of the distribution and \(v_i = \sum_{j=1}^{K} p_{i,j} * (j - m_i)^2\) is the variance of the distribution.

In addition to being more effective than other losses on the age estimation task, mean-variance loss also satisfies our needs to learn a relatively concentrated age distribution while capturing the age continuity for the adjacent aging bases.
The supervised age loss is formulated as below:

\begin{equation}
\mathbf{L}_{real} = L_{mv}(C(E(x)), y),
\label{real_age_loss}
\end{equation}

For guiding face aging, we apply the embedded age estimator at both the transformed identity encoding level and the generated image level (as shown in \textbf{Fig.} \ref{method_diag}).

\begin{equation}
\begin{split}
\mathbf{L}_{fake} &= \lambda_{fake1} L_{mv}(\hat{C}(PAT(E(x), t)), t) + \\
                                    & \lambda_{fake2} L_{mv}(\hat{C}(\hat{E}(G(PAT(E(x), t)))), t),
\label{fake_age_loss}
\end{split}
\end{equation}

When the age estimator \(\mathbf{\hat{C}}\) and encoder \(\mathbf{\hat{E}}\) are used on the transformed identity encodings and fake images, their weights are not updated during backpropagation.

\textbf{L1 Reconstruction Loss}
Another important aspect is to preserve the identity of the individual. We apply \(L1\) pixel-wise reconstruction loss on the synthesized face by setting the target age to its self-estimated age. Specifically, it is formulated as below:

\begin{equation}
\mathbf{L}_{idt} = \frac{1}{N} \sum_{i}^{N} || G(PAT(E(x_i), m_i)) - x_i ||_1,
\label{idt_loss}
\end{equation}

We have also experimented with a cycle-consistency loss as proposed in StarGAN \cite{choi2018stargan} to enforce the identity criteria but found that the pixel-wise \(L1\) reconstruction loss is sufficient to achieve the goal without extensive efforts in tuning the hyper-parameters.

\begin{figure*}[t!]
    
    \includegraphics[width=\textwidth]{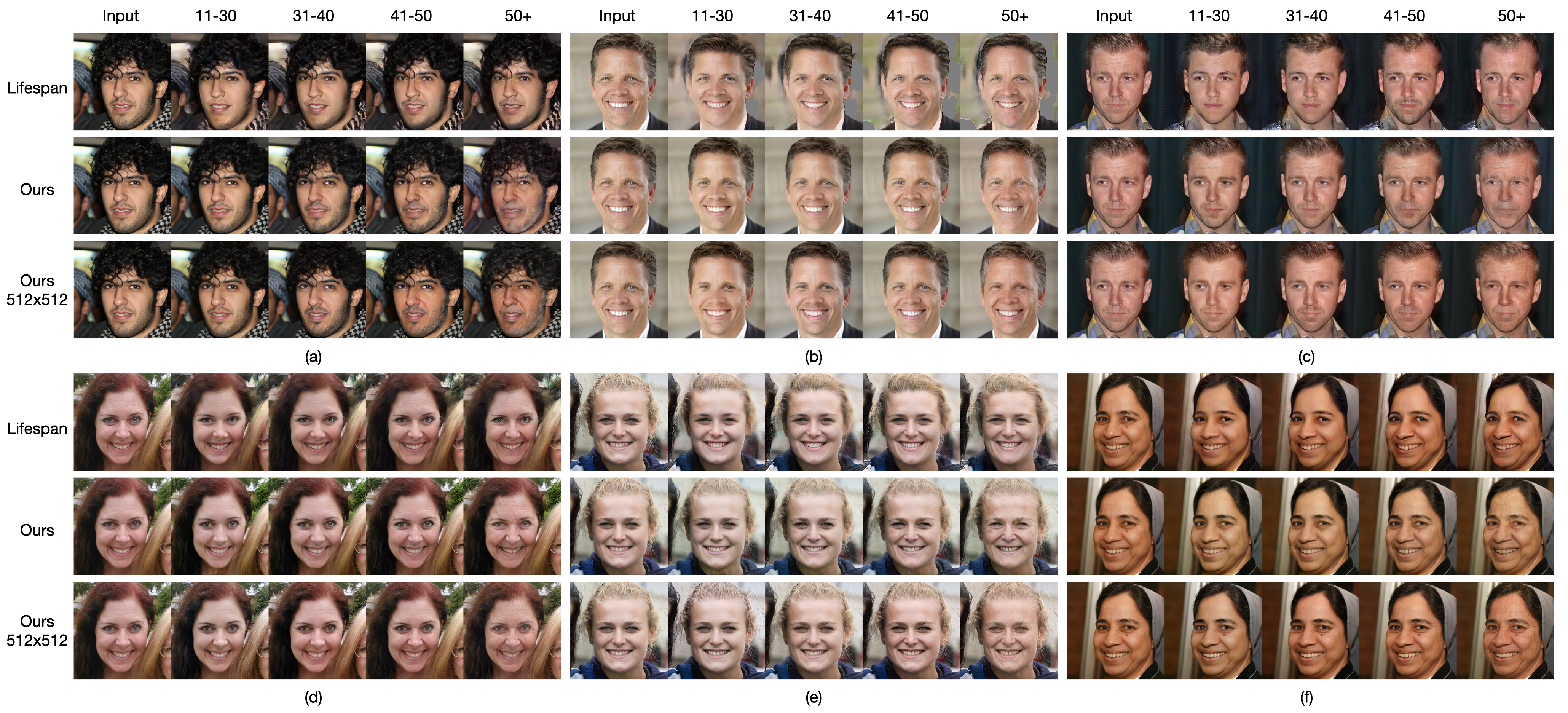}

    \caption{Comparisons on FFHQ \cite{karras2019style} among Lifespan\cite{orel2020lifespan}, ours and ours (512x512). Lifespan does not have an explicit age group at 11-30 and 41-50 so images for these 2 groups are generated using linear interpolation between 2 neighboring anchor classes. As shown, our generated images provide more aging details, such as skin wrinkles and color of the beard, on different parts of the face. In the example (f) in particular, both of our models well preserve her personal traits (a mole), comparing to the Lifespan model.}
    \label{ffhq_img}
\end{figure*}

\begin{figure*}[t!]
\centering
	\vspace{-0.3cm}
    \includegraphics[width=0.98\textwidth]{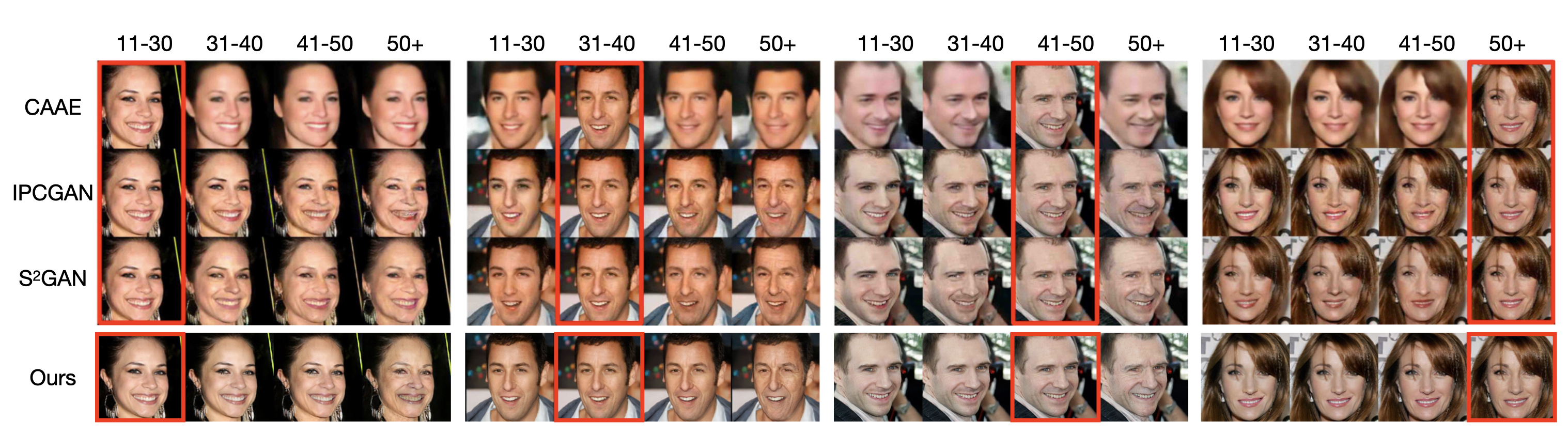}
    \caption{Comparisons on CACD2000 \cite{chen2014cross} among CAAE\cite{zhang2017age} IPCGAN \cite{wang2018face}, S\(^2\)GAN \cite{he2019s2gan} and ours. The input images are wrapped in red boxes.}
    \label{cacd_img}
    \vspace{-0.2cm}
\end{figure*}

\begin{figure*}[t!]
    \centering
    \includegraphics[width=0.99\textwidth]{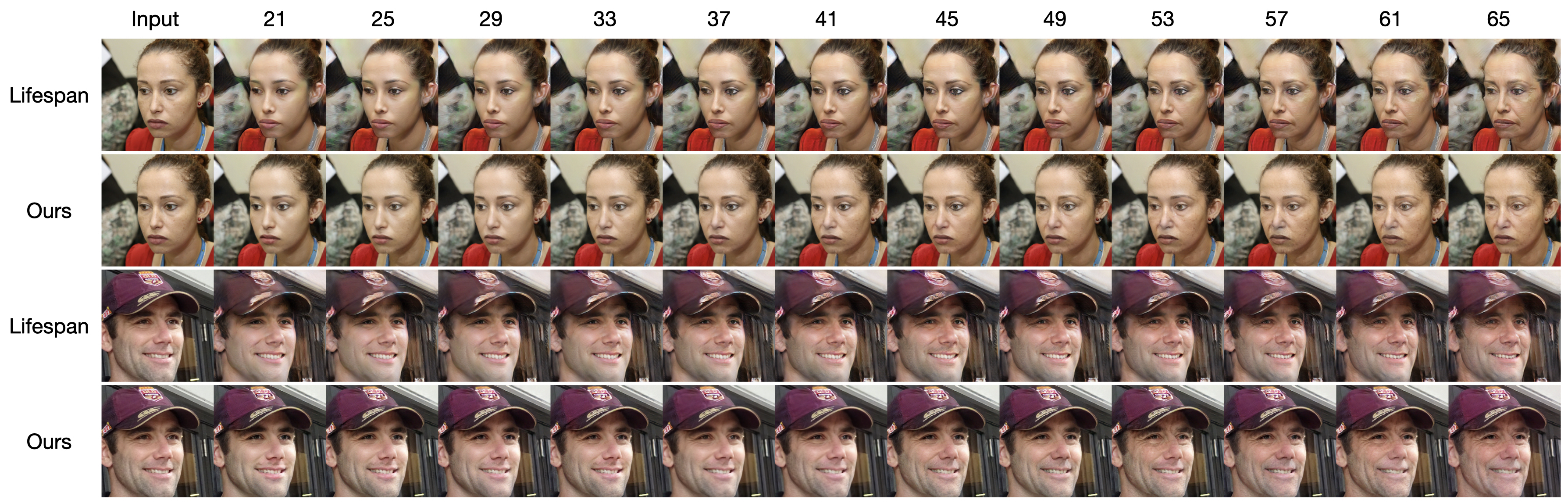}
    \caption{Continuous aging from 21 to 65. The age gap is chosen as 4 due to the limited space. As shown, the linear interpolation-based method used by Lifespan \cite{orel2020lifespan}, some personal traits are altered (such as mouth shape, beard, hats). Further, our method generates more realistic aging effects with minimal artifacts. Continuous aging of 1-year incremental change in Supplementary.}
    \label{continuous_img}
\end{figure*}
\textbf{Adversarial Loss}
To produce high fidelity images, we apply GAN loss in the unconditional adversarial training manner. More specifically, we adopt PatchGAN \cite{pix2pix2017} discriminator and optimize on the hinge loss, formulated as the following:

\begin{equation}
\begin{split}
\mathbf{L}_{adv-D} &= E_{z \sim p_{data}(z)}[max(1 - D(z), 0)] + \\
                                    &E_{(x,t) \sim p_{data}(x)}[max(1 + D(G(PAT(E(x), t))), 0)],
\label{adv_d_loss}
\end{split}
\end{equation}
where we denote the data distribution as $x \sim p_{data}(x)$ and $z \sim p_{data}(z)$. 

\begin{equation}
\mathbf{L}_{adv-G} = E_{(x,t) \sim p_{data}(x)}[-D(G(PAT(E(x), t)))],
\label{adv_g_loss}
\end{equation}

In the experiment, we observe that sampling real examples of the age equal or close to the target age \(t_i\) for training the discriminator helps to stabilize the learning process. 

All objectives are optimized jointly with different balancing coefficients as the following:

\begin{equation}
\underset{E, C, PAT, G}{\text{min}} \lambda_{age} (L_{real} + L_{fake}) + \lambda_{idt} L_{idt} + \lambda_{adv} L_{adv-G},
\label{final_g_loss}
\end{equation}

\begin{equation}
\underset{D}{\text{min}} L_{adv-D},
\label{final_d_loss}
\end{equation}

\section{Experiments}

\textbf{Datasets}
We evaluated our model on FFHQ \cite{karras2019style} and CACD2000 \cite{chen2014cross}. FFHQ includes 70000 images with 1024x1024 resolution. Following the data preprocessing procedures as \cite{orel2020lifespan}, we take images with id 0-68999 as the training set and 69000-69999 for testing and filter out images with low confidence in differentiating the gender, low confidence in estimating the age, wearing dark glasses, extreme pose, and angle based on the facial attributes annotated by Face++\footnote{Face++ facial attribute annotation API: \url{https://www.faceplusplus.com/}}. As the annotation from \cite{orel2020lifespan} only includes the age group label, we acquire the age label information from \cite{yao2020high}. To reconcile both age group labels and age labels, we further filter out images in which the age label disagrees with the age group label. This results in 12488 male and 13563 female images for training, and 279 male and 379 female images for testing. CACD2000 consists of 163446 images where age ranges from 14 to 62 years old. We randomly take 10\% of data for evaluation. We use Face++ to separate the images into male and female and extract the facial landmarks using Dlib\footnote{Dlib toolkit: \url{http://dlib.net/}}.

\textbf{Implementation}
Since aging patterns are different between males and females, we train two separate models on the FFHQ dataset for both 256x256 and 512x512 resolutions. Model architecture is modified based on CycleGAN \cite{zhu2017unpaired}. Please refer to the Supplementary for the detailed model architecture and optimization settings. $\lambda_{mv1}$ and $\lambda_{mv2}$ are set to 0.05 and 0.005 in \textbf{Eq.} (\ref{mean_variance_loss}). $\lambda_{fake1}$ and $\lambda_{fake2}$ are set to 0.4 and 1 in \textbf{Eq.} (\ref{fake_age_loss}). In \textbf{Eq.} (\ref{final_g_loss}), $\lambda_{age}$, $\lambda_{idt}$ , and $\lambda_{adv}$ are set to 0.05, 1, and 1 respectively.

\subsection{Qualitative Evaluation}

\textbf{Face Aging}
We present our test results on FFHQ, comparing with results from \cite{orel2020lifespan}. Images for \cite{orel2020lifespan} are generated using their provided code\footnote{Lifespan official code: \url{https://github.com/royorel/Lifespan_Age_Transformation_Synthesis}}. To illustrate the model performance across different ages, we show 6 input examples from 4 representative age groups (\(<\)30, 30-40, 40-50, 50+) and generate the results for each group. The target ages for our model are chosen as 25, 35, 45, and 55 respectively. As can be seen in \textbf{Fig.} \ref{ffhq_img}, the images generated by our model result in fewer artifacts and exhibit more clear aging details, such as beard color change (example a,c) and wrinkles on different parts of the face (see example b,c,d,e). A convincing detail in example (f) shows that the personal traits (a mole) are well preserved using our models.

We also directly generates images on CACD2000 using the models trained on FFHQ in the resolution of 256x256 to compare with CAAE\cite{zhang2017age}, IPCGAN \cite{wang2018face}, and S\(^2\)GAN \cite{he2019s2gan} in \textbf{Fig.} \ref{cacd_img}. The demonstrated images are the presented examples in \cite{he2019attgan}, which is the state-of-the-art work on CACD2000. For all age groups, our model presents more evident and fine-grained aging effects comparing with all previous works.

\begin{figure}[h]
\centering
\vspace{-0.2cm}
\includegraphics[width=0.45\textwidth]{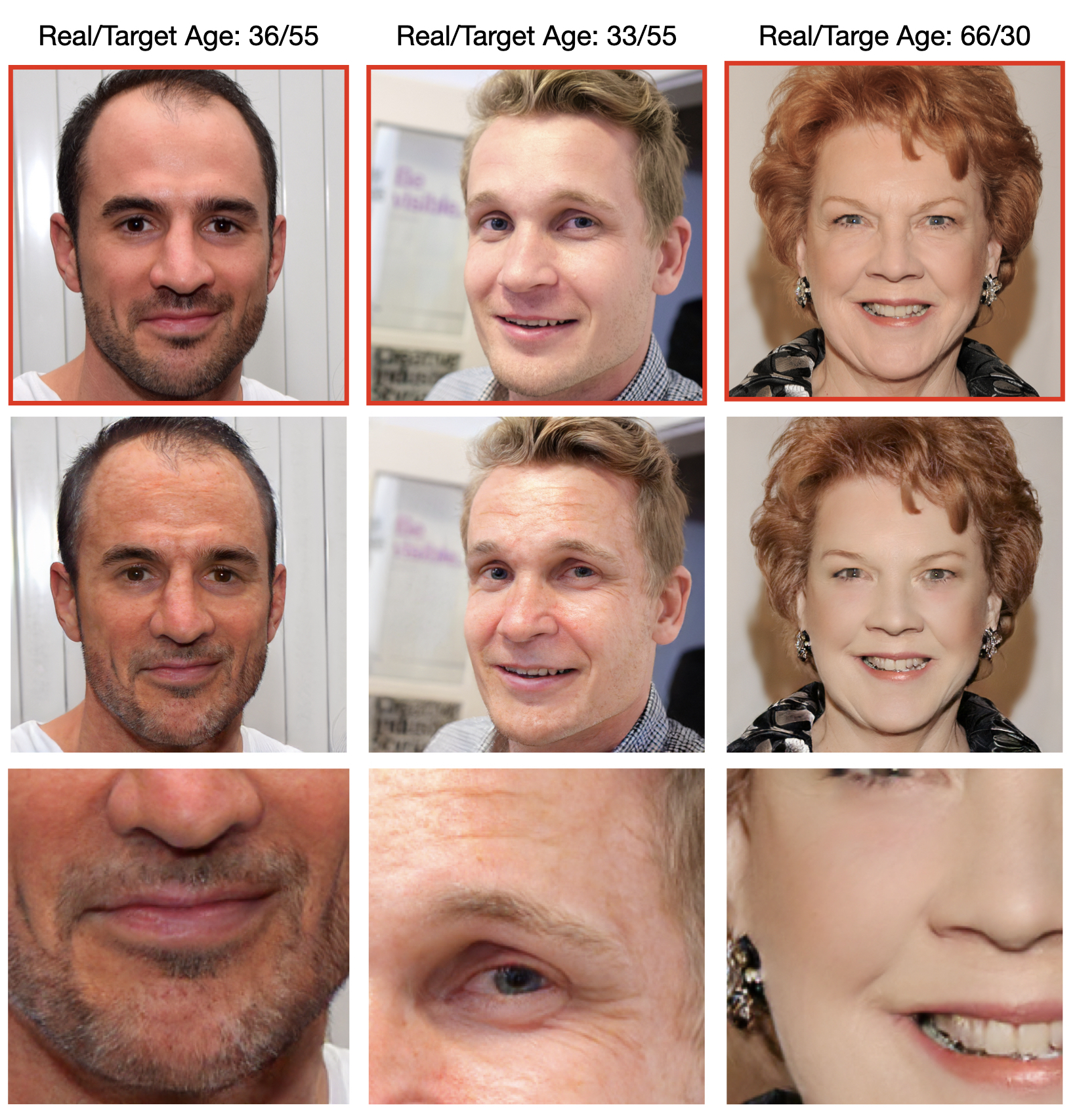}
\caption{Aging details: enlarged face crops to show details for beard, wrinkle, and skin smoothness. Input images are in red boxes.}
\label{details_img}
\vspace{-0.2cm}
\end{figure}

\textbf{Continuous Aging}
In \textbf{Fig.} \ref{continuous_img}, we illustrate some examples of continuous aging results comparing with \cite{orel2020lifespan}. We choose an age step of 4 to present due to the limited space. A gradual and smooth natural aging process (e.g. wrinkle depth change, beard, pigmentation on face) can be observed from our images while retaining personal traits. The interpolation-based method in Lifespan, however, lacks the ability to generate images of well-aligned target ages and does not preserve certain personalized information.

\textbf{Aging Details} Here, we show that the generated images express a significant level of aging details on different parts of the face. In \textbf{Fig.} \ref{details_img}, we demonstrate three enlarged face crops from the generated images, which give a clear and detailed view of enhanced wrinkles, skin smoothness, color change of beard and eyebrow.

\subsection{Quantitative Evaluation}
\textbf{Identity Preservation}
To evaluate identity preservation, we adopt the face verification rate metric. Specifically, we followed the evaluation protocol of \cite{he2019s2gan} on an age group basis for a fair comparison with prior works. We calculate the face verification rate between all combination of image pairs, i.e. (test, 10-29), (test, 30-39), ...,(30-39, 40-49), (40-49, 50-59). Face verification score is obtained from Face++ and the threshold is set as 76.5 (@FAR=1e-5). The complete results are presented in \textbf{Table} \ref{table:idt_cacd_table} and \ref{table:idt_ffhq_table} for CACD2000 and FFHQ respectively. As the results suggest, our model achieves the highest face verification rate for both datasets among all candidates, which indicates it best meets the identity preservation requirement of the task.

\begin{figure*}[t!]
\centering
\vspace{-0.2cm}
\includegraphics[width=0.98\textwidth]{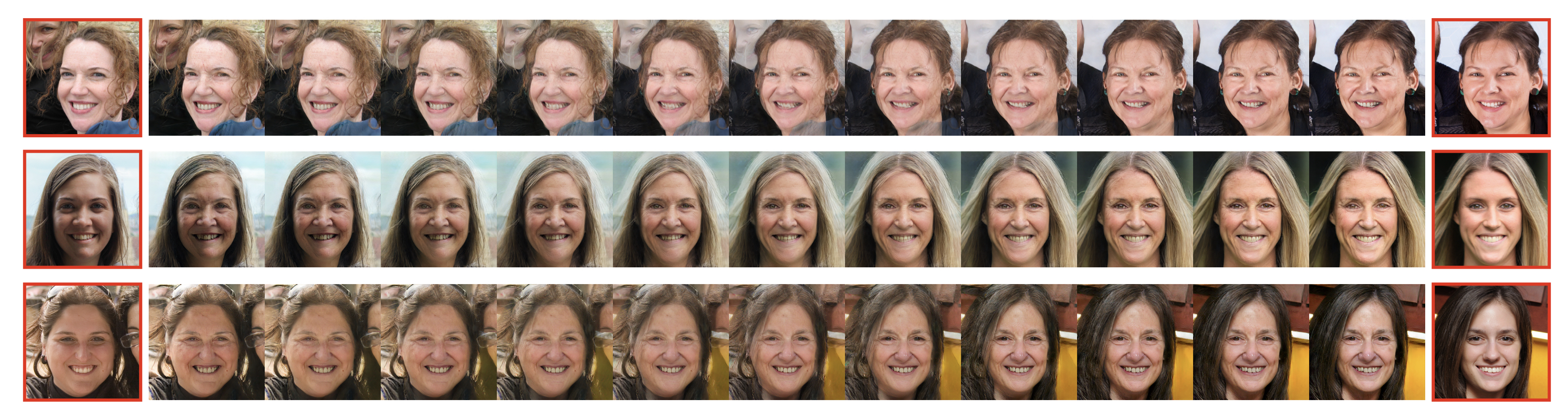}
\caption{Linear interpolation between transformed identity encodings. Real images are in red boxes. From left to right, we linearly interpolate between two images' transformed identity encodings at the same target age 65. Personal traits, such as eye color and teeth shape, smoothly change from one person to the other.}
\label{mix_identity}
\vspace{-0.4cm}
\end{figure*}

\begin{table}[h!]
  \centering
  \begin{tabular}{cc}
    \toprule
    & Average of All Pairs \\
    \midrule
    CAAE \cite{zhang2017age} & 60.88\% \\
    IPCGAN \cite{wang2018face} & 91.40\% \\
    S\(^2\)GAN \cite{he2019s2gan} & 98.91\% \\
    Lifespan \cite{orel2020lifespan} & 93.25\% \\
    Ours & \boldmath{$99.97\%$} \\
    \bottomrule
  \end{tabular}
  \caption{Evaluation of identity preservation in terms of face verification rates on CACD2000 \cite{chen2014cross}. Pair-wise results are presented in Supplementary.}
    \label{table:idt_cacd_table}
    \vspace{-0.4cm}
\end{table}

\begin{table}[h!]
  \centering
  \begin{tabular}{cc}
    \toprule
    & Average of All Pairs \\
    \midrule
    Lifespan \cite{orel2020lifespan} & 87.11\% \\
    Ours & \boldmath{$99.98\%$} \\
    \bottomrule
  \end{tabular}
  \caption{Evaluation of identity preservation in terms of face verification rates on FFHQ \cite{karras2019style}. Pair-wise results are presented in Supplementary.}
    \label{table:idt_ffhq_table}
    \vspace{-0.1cm}
\end{table}

\textbf{Aging Accuracy}
In terms of assessing aging accuracy, we use an unbiased age estimator to infer the age of the generated images. To be able to compare with prior group-based methods on CACD2000, we generate our images aligning with their age group settings in which we adaptively increment/decrement by a factor of 10 (age group size) from input image's real age as the target age for generation, i.e. target age 33 is used for generating an image of age group 30-40 given current age of 23. As we neither have the access to \cite{he2019s2gan}'s evaluation age estimator nor their pre-trained model for assessing our model and doing a direct comparison, we instead use Face++'s age estimation results on our model and one of accessible prior work IPCGAN \cite{wang2018face}, which is also evaluated in \cite{he2019s2gan} to show relative comparison. Evaluation of FFHQ follows the same procedure as CACD2000. The evaluation results are shown in \textbf{Table} \ref{table:age_cacd_table} and \ref{table:age_ffhq_table} for CACD2000 and FFHQ respectively. As the results suggest, our model evaluated using Face++ has a more reasonable mean age at each age group than IPCGAN \cite{wang2018face} and Lifespan \cite{orel2020lifespan} on CACD2000 and has a similar performance as Lifespan on FFHQ.

\begin{table}[t!]
  \centering
  \begin{tabular}{ccccc}
    \toprule
    & 10-29 & 30-39 & 40-49 & 50+ \\
    \midrule
    CAAE \cite{zhang2017age} & 29.6 & 33.6 & 37.9 & 41.9 \\
    S\(^2\)GAN \cite{he2019s2gan} & 24.0 & 36.0 & 45.7 & 55.3 \\
    IPCGAN \cite{wang2018face} & 27.4 & 36.2 & 44.7 & 52.5 \\
    \midrule
    IPCGAN \cite{wang2018face} (Face++) & 42.4 & 47.1 & 51.9 & 56.0 \\
    Lifespan \cite{orel2020lifespan} (Face++) & - & 40.2 & - & 64.3 \\
    Ours (Face++) & 30.5 & 38.7 & 46.9 & 60.0 \\
    \bottomrule
  \end{tabular}
  \caption{Comparison of the mean age of generated images in each age group evaluated using Face++ on CACD2000 \cite{chen2014cross}.}
    \label{table:age_cacd_table}
\end{table}

\begin{table}[t!]
  \centering
  \begin{tabular}{ccccc}
    \toprule
    & 10-29 & 30-39 & 40-49 & 50+ \\
    \midrule
    Lifespan \cite{orel2020lifespan} & - & 38.4 & - & 63.8 \\
    Ours & 30.7 & 38.4 & 47.7 & 62.1 \\
    \bottomrule
  \end{tabular}
  \caption{Comparison of the mean age of generated images in each age group evaluated using Face++ on FFHQ \cite{karras2019style}.}
    \label{table:age_ffhq_table}
\end{table}
\textbf{Image Fidelity}
Considering the image fidelity, we adopt the Fréchet Inception Distance (FID) \cite{heusel2017gans} metric to evaluate our model. Similar to the image generation settings as before, we calculated the FID on the generated images corresponding to the same age group as theirs on CACD2000. For comparing with \cite{orel2020lifespan} on FFHQ, we calculate the FID on the generated images, that share the same age group range. The results are shown in the \textbf{Table} \ref{table:fid}. On both datasets, our model achieves the lowest FID, which quantitatively demonstrates superiority in the image quality aspect.

\begin{table}[t!]
  \centering
  \begin{tabular}{ccc}
    \toprule
    & CACD2000 & FFHQ \\
    \midrule
    CAAE \cite{zhang2017age} & 44.2 & - \\
    IPCGAN \cite{wang2018face} & 9.1 & - \\
    S\(^2\)GAN \cite{he2019s2gan} & 8.4 & - \\
    Lifespan \cite{orel2020lifespan} & 11.7 & 26.2 \\
    Ours & \boldmath{$6.7$} & \boldmath{$18.5$} \\
    \bottomrule
  \end{tabular}
  \caption{FID evaluation: lower is better.}
    \label{table:fid}
    \vspace{-0.5cm}
\end{table}

\subsection{Model Interpretability and Ablation Study}
\begin{figure}
\centering
\includegraphics[width=0.46\textwidth]{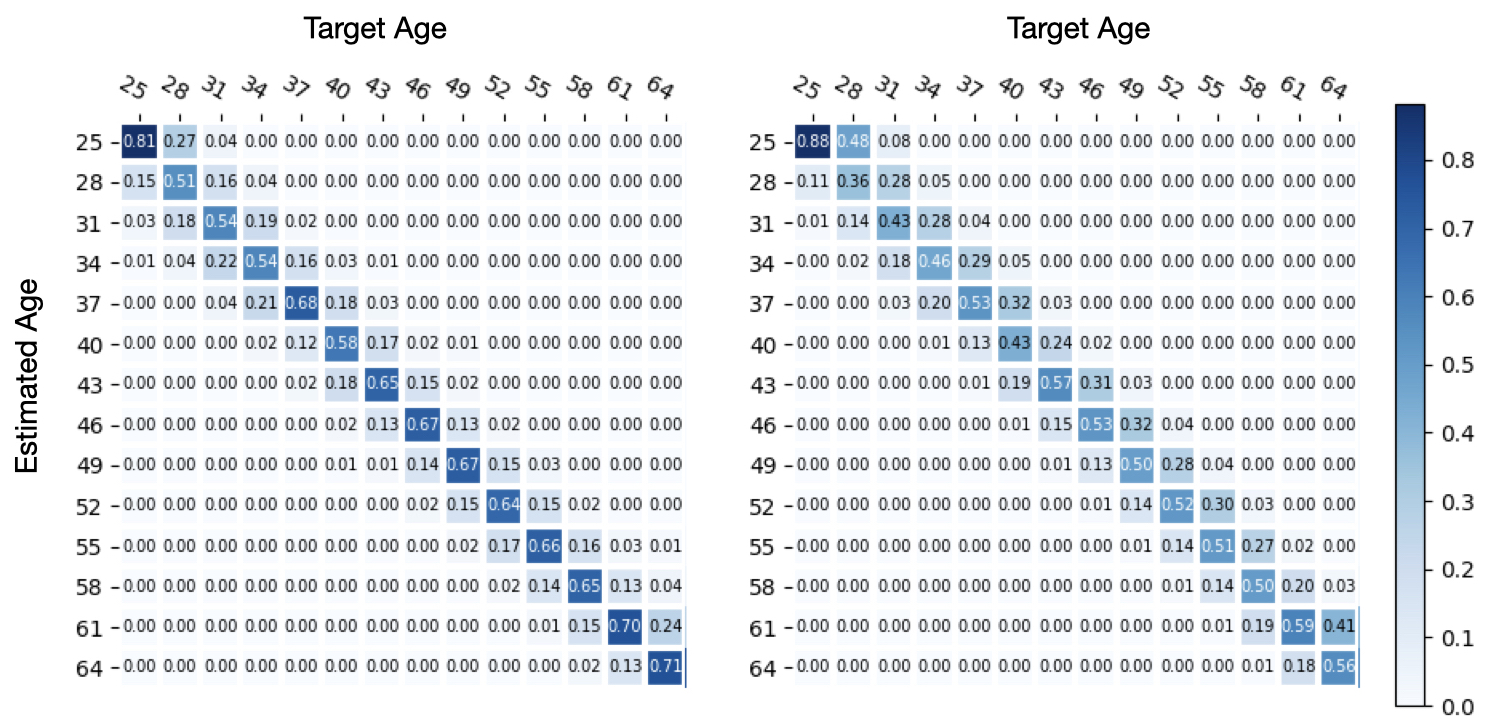}
\caption{Confusion matrices of continuous face aging. Left: age estimation on the self-estimated aging embeddings (proposed). Right: age estimation on the linear interpolated aging embeddings. The age step is chosen as 3 based on the  MAE of the estimator.} 
\label{age_matrix}
\vspace{-0.5cm}
\end{figure}

\textbf{Continuous Aging}
To evaluate how well our model generates synthesized images in a continuous setting, we use an age estimator to predict age on the generated fake images from 25 to 65 of our approach and the linear interpolation approach performed between anchor aging bases. The anchor basis is generated by taking the mean of every aging bases within an age group. We calculate a confusion matrix in terms of aging accuracy for each approach using the age estimator jointly trained on the FFHQ dataset. \textbf{Fig.}  \ref{age_matrix} indicates that our generated fake images express a more evident continuous aging trend with much higher aging accuracy than the linear interpolation approach.

\textbf{Interpolation between Two Identities in Latent Space} In \textbf{Fig.}  \ref{mix_identity}, we further illustrate that our proposed model also learns a disentangled representation of age and identity in latent space. We linearly interpolate between the two transformed identity encodings of the same age and different identities and then generate images for the interpolated encodings. As shown in the figure, the identity changes gradually while maintaining the respective age.

\textbf{Use of the Residual Embedding} One of the key innovative design of our model architecture is the formulation of the personalized age embedding, which incorporates both personalized aging features of the individual and shared aging effects among the entire population. To better illustrate and understand the effectiveness of the design, we train a model without adding the residual embedding (i.e. directly applying the target age's exemplar-face aging basis \(a_{i,j=t_i}\)), and compare with the proposed method.

\begin{figure}[h!]
    \centering
    \vspace{-0.2cm}
    \includegraphics[width=0.5\textwidth]{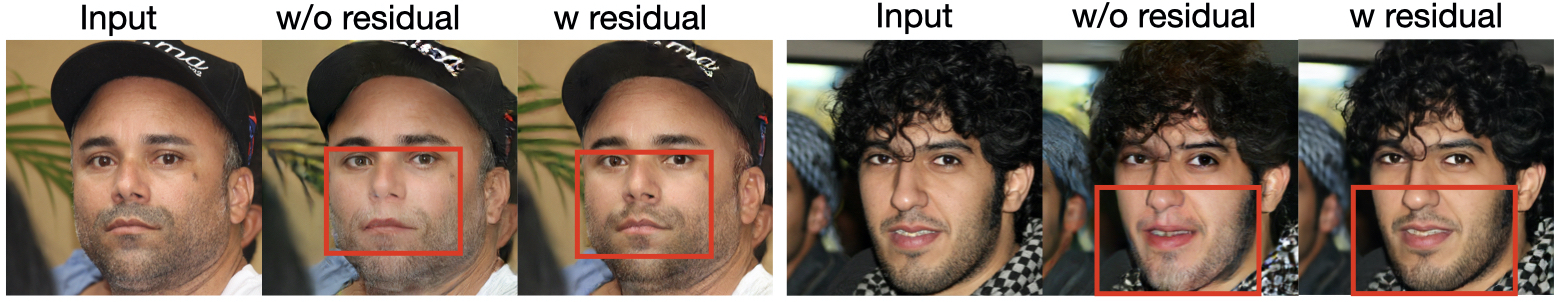}
    \includegraphics[width=0.48\textwidth]{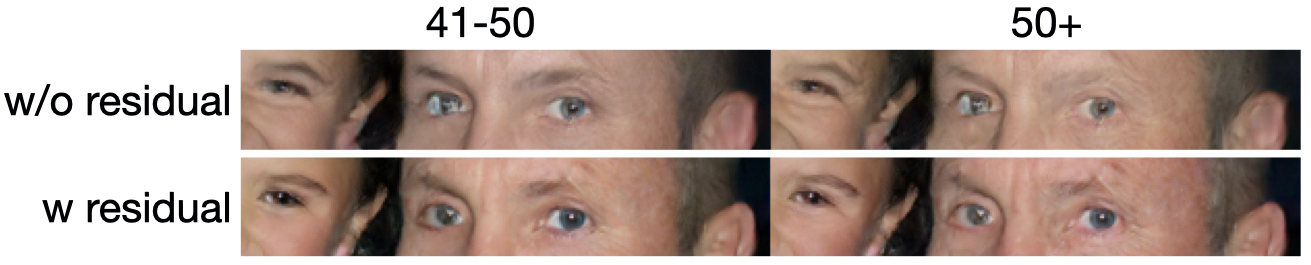}
    \caption{Enlarged ablation examples. Top-row target age: 11-30. The nose shape and beard were not preserved in the w/o residual example (top). Artifacts in eyes are commonly seen in older age groups w/o residual (bottom).}
    \vspace{-0.5cm}
    \label{ablation_img}
\end{figure}

In \textbf{Fig.} \ref{ablation_img}, we display a few examples with highlighted/enlarged regions comparing results w/wo residual embeddings. Noticeably, more unnatural artifacts and a tendency to examplar-face modification are observed in the images generated without residual embeddings.

\section{Conclusions}
In this work, we introduce a novel approach to the task of face aging with a specific focus on the continuous aging aspect. We propose a unified framework to learn continuous aging bases via introducing an age estimation module to a GAN-based generator. The designed \(\mathbf{PAT}\) module further enhances the personalization of the exemplar-face aging bases, which results in more natural and realistic generated face images overall. The experiments qualitatively and quantitatively show superior performance on the aging accuracy, identity preservation, and image fidelity on two datasets compared to prior works. Furthermore, the proposed network structure can also be applied to other multi-class domain transfer tasks to avoid group-based training and achieve a more accurate continuous modeling.

\clearpage
{\small
\bibliographystyle{ieee_fullname}
\bibliography{egbib}
}

\clearpage

\section{Supplementary}
\subsection{Network Architecture and Optimization Settings}

During training, we use Adam optimizer with the learning rate of 0.0002 and batch size of 20 and 5 for 256 and 512 model respectively. The model is trained for 200 epochs and learning rate is linearly decayed over last 100 epochs.
\begin{table}[h]
  \centering
  \begin{tabular}{ccccc}
    \toprule
    Layer & Stride & Act. & Norm & Output Shape \\
    \midrule
    Input & - & - & - & 256x256x3 \\
    \midrule
    Conv. 7 x 7 & 1 & ReLU & Spectral & 256x256x64 \\
    Conv. 3 x 3 & 2 & ReLU & Spectral & 128x128x128 \\
    Conv. 3 x 3 & 2 & ReLU & Spectral & 64x64x256 \\
    \midrule
    Res. Block & 1 & ReLU & Spectral & 64x64x256 \\
    Res. Block & 1 & ReLU & Spectral & 64x64x256 \\
    Res. Block & 1 & ReLU & Spectral & 64x64x256 \\
    Res. Block & 1 & ReLU & Spectral & 64x64x256 \\
    Res. Block & 1 & ReLU & Spectral & 64x64x256 \\
    Res. Block & 1 & ReLU & Spectral & 64x64x256 \\
    \bottomrule
  \end{tabular}
  \caption{Identity Encoder $\mathbf{E}$ specification. Spectral means spectral normalization \cite{miyato2018spectral} is applied after each convolutional layer.}
  
    \label{table:encoder}
\end{table}

\begin{table}[h]
  \centering
  \begin{tabular}{ccccc}
    \toprule
    Layer & Stride & Act. & Norm & Output Shape \\
    \midrule
    Encoding & - & - & - & 64x64x256 \\
    \midrule
    Res. Block & 1 & ReLU & Instance & 64x64x256 \\
    Res. Block & 1 & ReLU & Instance & 64x64x256 \\
    Res. Block & 1 & ReLU & Instance & 64x64x256 \\
    \midrule
    Deconv. 3 x 3 & 2 & ReLU & Instance & 128x128x128 \\
    Deconv. 3 x 3 & 2 & ReLU & Instance & 256x256x64 \\
    \midrule
    Conv. 7 x 7 & 1 & Tanh & - & 256x256x3 \\
    \bottomrule
  \end{tabular}
  \caption{Generator $\mathbf{G}$ specification. Instance means instance normalization \cite{ulyanov2016instance} is applied after each convolutional layer.}
    \label{table:decoder}
\end{table}

\begin{table}[h]
  \centering
  \begin{tabular}{ccc}
    \toprule
    Layer & Norm & Output Shape \\
    \midrule
    Encoding & - & 64x64x256 \\
    \midrule
    GAP & - & 1x1x256 \\
    Flatten & - & 256 \\
    Linear & Weight & 100 \\
    \bottomrule
  \end{tabular}
  \caption{Age estimator $\mathbf{C}$ specification. GAP means global average pooling. Weight means weight normalization \cite{salimans2016weight} is applied to linear layer (bias term are set to zero).}
    \label{table:estimator}
\end{table}

Detailed network architectures for $\mathbf{E}$, $\mathbf{G}$ and $\mathbf{C}$ are presented in \textbf{Table} \ref{table:encoder}, \ref{table:decoder} and \ref{table:estimator} respectively. 

\subsection{Pair-wise Identity Preservation Results}
Here, we provide the complete pair-wise identity preservation comparison using Face++ in \textbf{Table} \ref{table:idt_cacd_full} and \ref{table:idt_ffhq_full} for CACD2000 \cite{chen2014cross} and FFHQ \cite{karras2019style}, respectively. As can be seen, our model achieves the highest verification rate in every aspects compared to prior works.

\subsection{More Aging Results}
\begin{figure*}[h!]
    \centering
    \vspace{-0.8cm}
    \includegraphics[width=0.95\textwidth]{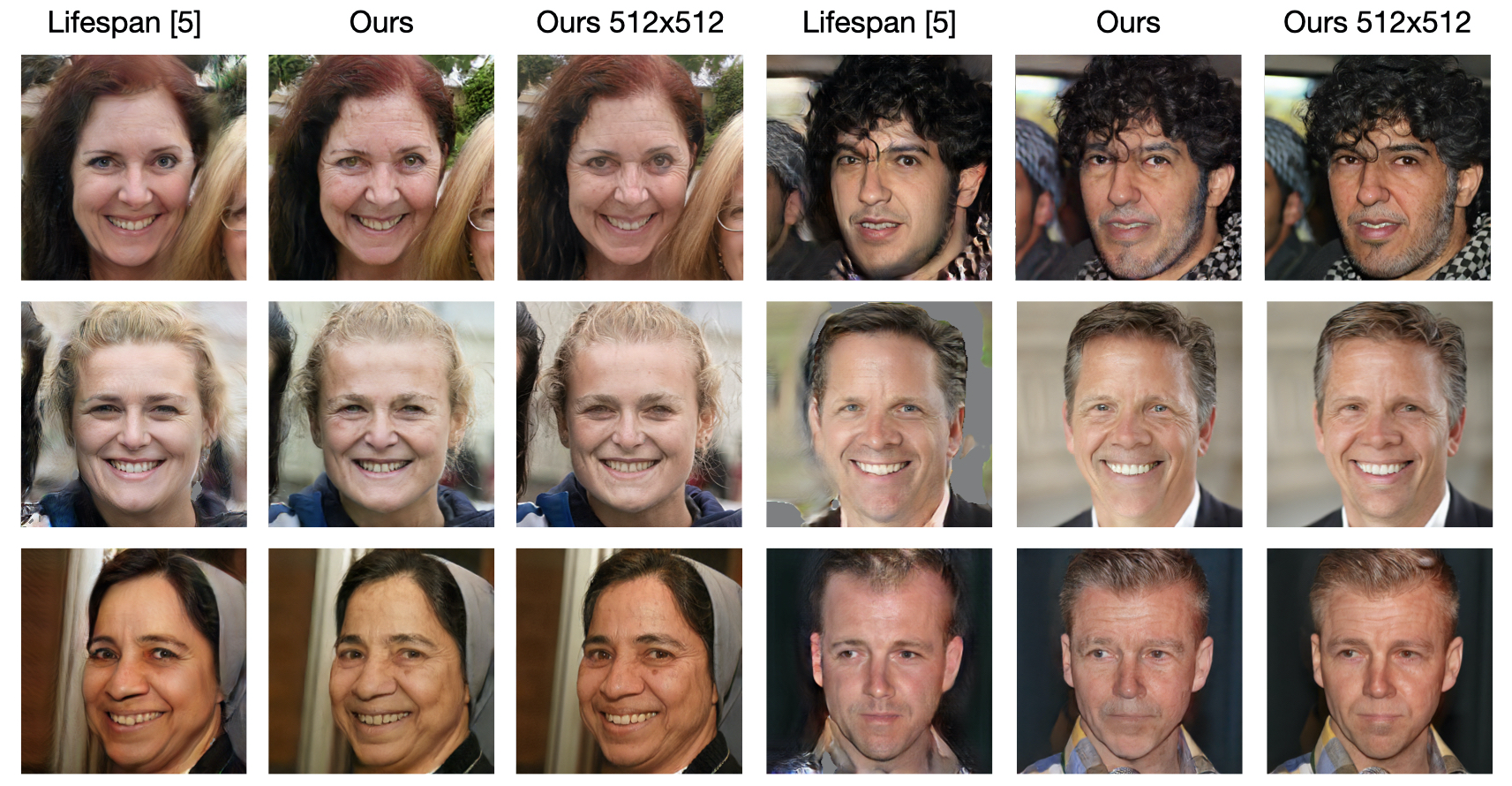}
    \caption{Enlarged examples of generated age 50+. Our model demonstrate better details in aging effects such as wrinkles and beard change.}
    \label{fig:enlarged_50s}
\end{figure*}
\begin{figure*}[t!]
\centering
    \includegraphics[width=0.98\textwidth]{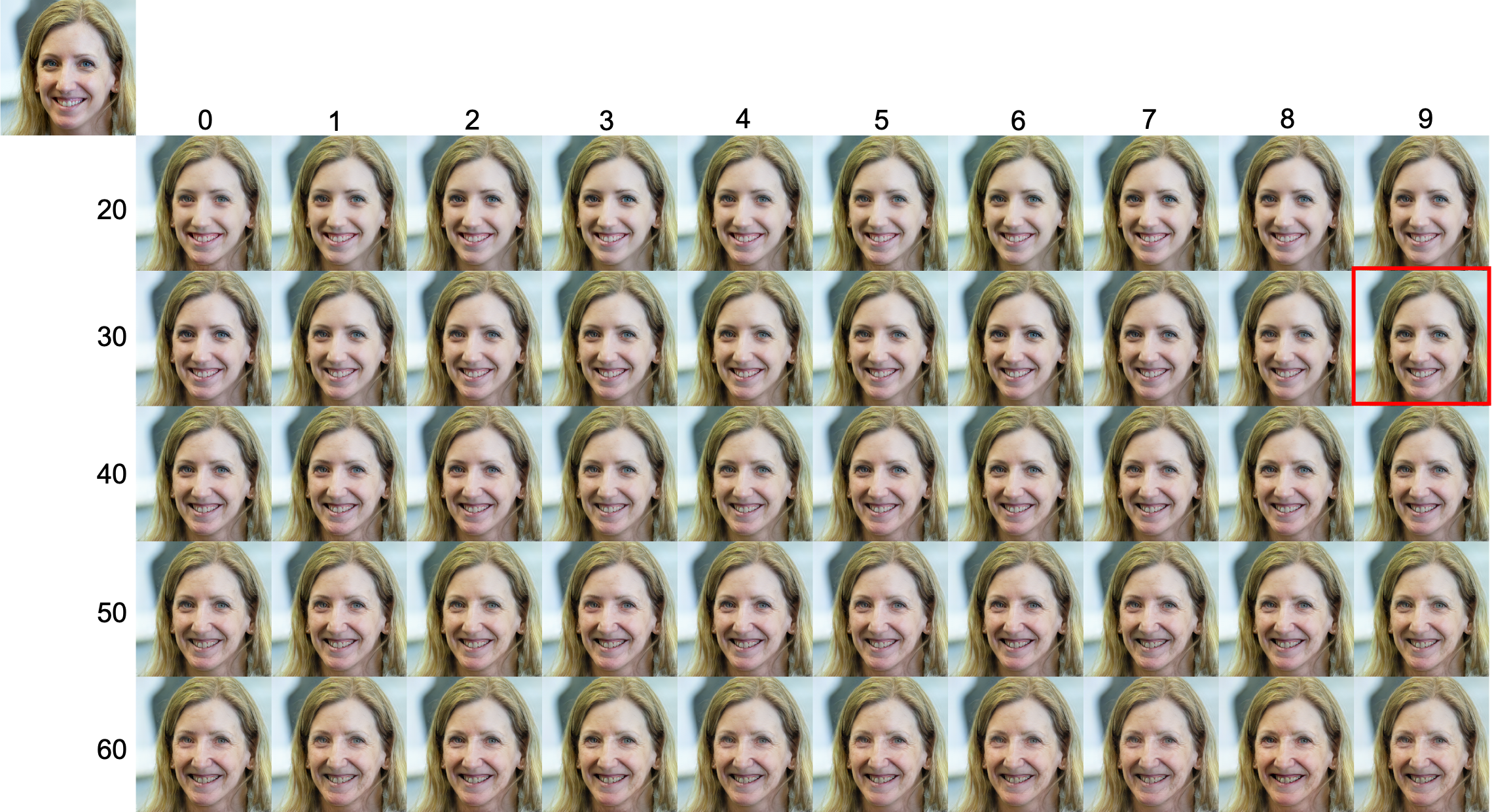}
	\includegraphics[width=0.98\textwidth]{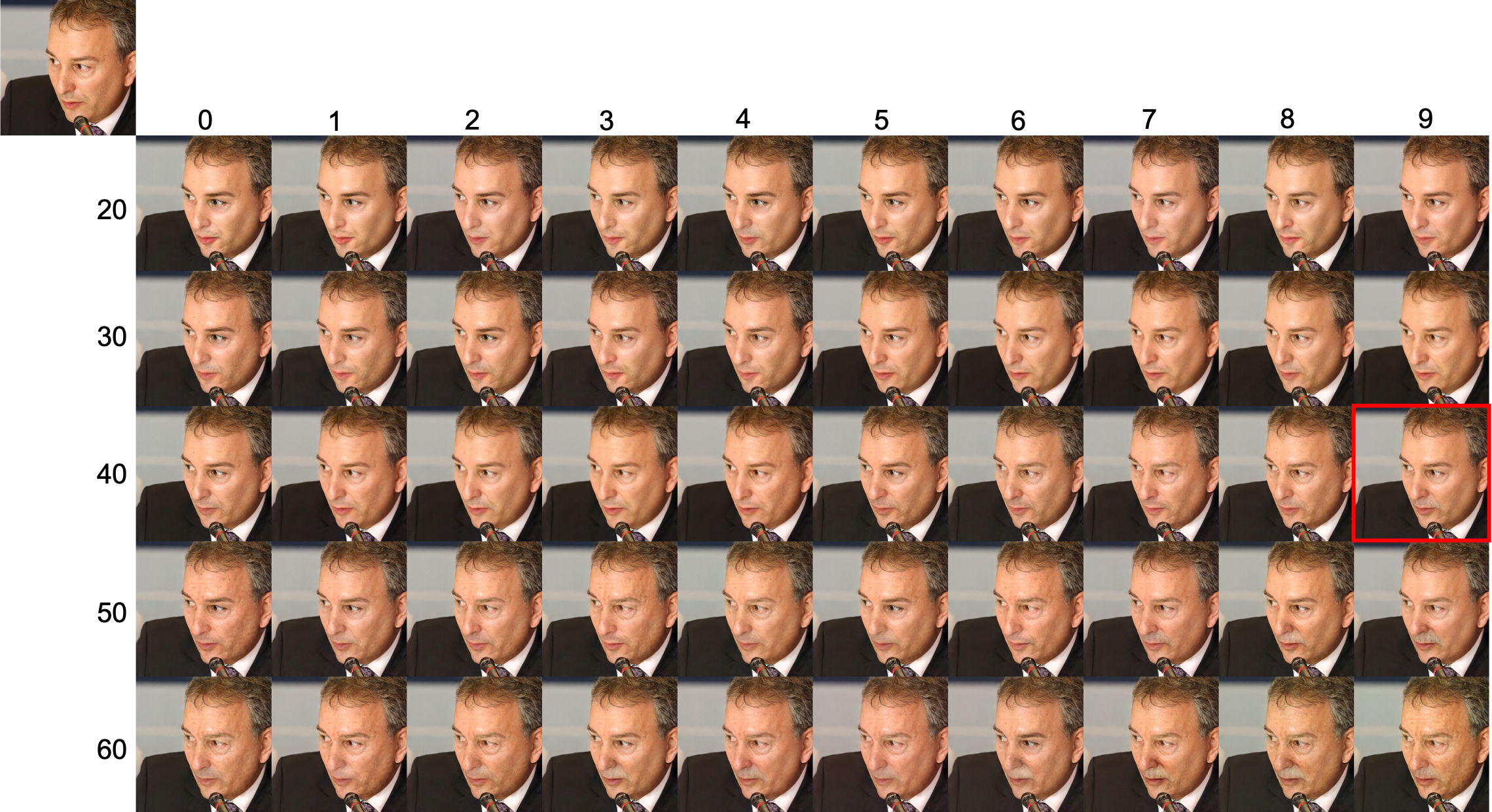}
    \caption{Complete continuous aging results from age 20 to 69. Input is at the top left corner of each image grid. Generated image of real age is in the red box.}
    \label{continuous_full_img}
    
\end{figure*}

\textbf{Continuous Aging.} We generate the complete continuous aging results of a person from age 20 to age 69 and the results are displayed in \textbf{Fig.} \ref{continuous_full_img}. As shown, aging proceeds in a natural and gradual manner.

\textbf{Enlarged Comparison of group 50+.} In \textbf{Fig.} \ref{fig:enlarged_50s}, we show the enlarged generated images of age group 50+. Our model is able to generate fine aging details aligned with the target age group.

\subsection{Limitations}
\begin{figure*}[h]
    \centering
    \vspace{-0.2cm}
    \includegraphics[width=0.95\textwidth]{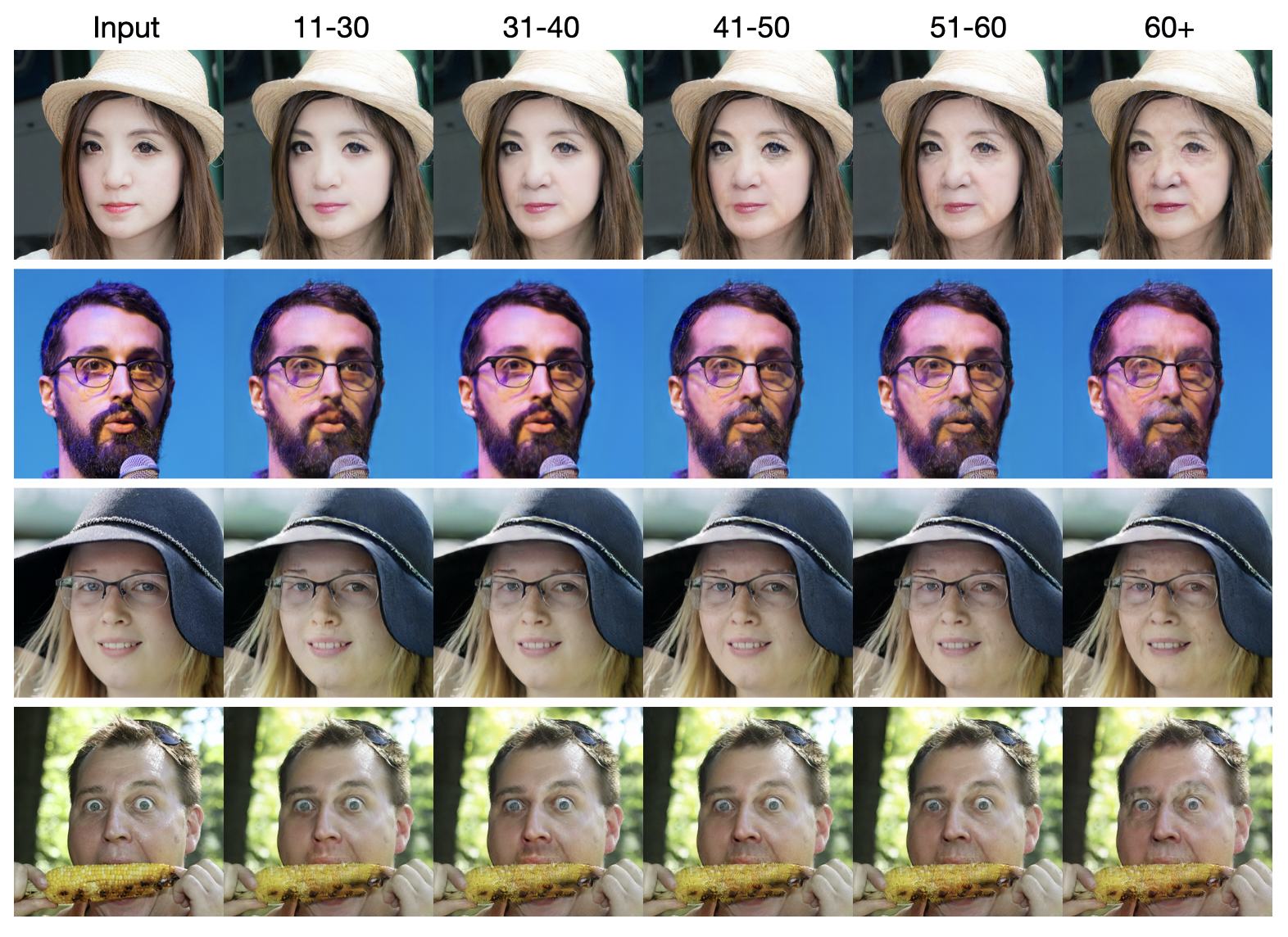}
    \caption{Failure cases: heavy makeup with hat, glasses with bad lighting. Our model could not best capture the personalized information such as skin texture in these cases. }
    \vspace{-0.3cm}
    \label{fig:failure_cases}
\end{figure*}

While our work can generate natural face aging, we also observe some failure cases when generating outputs for input image with hats and glasses or faces with heavy makeups (in \textbf{Fig.} \ref{fig:failure_cases}). The model also does not work well for extreme target age like 95-year-old, where the corresponding exemplar-face aging basis is hardly trained due to lack of data for those minority classes.

\clearpage
\begin{table*}[htbp]
  \centering
    \vspace{-8cm}
    \vspace{4cm}
  \begin{tabular}{cccc}
    \toprule
    & Average of All Pairs & Hardest Pair & Easiest Pair \\
    \midrule
    CAAE \cite{zhang2017age} & 60.88\% & (test, 50+): 2.0\% & (40-49, 50+): 99.97\% \\
    IPCGAN \cite{wang2018face} & 91.40\% & (10-29, 50+): 62.98\% & (40-49, 50+): 99.98\% \\
    S\(^2\)GAN \cite{he2019s2gan} & 98.91\% & (10-29, 40-49): 94.08\% & (40-49, 50+): 99.96\% \\
    Lifespan \cite{orel2020lifespan} & 93.25\% & (test, 50-69): 80.94\% & (30-39, 50-69): 99.75\% \\
    Ours & \boldmath{$99.97\%$} & (test, 40-49): 99.96\% & (test, 30-39): 100.00\% \\
    \bottomrule
     
  \end{tabular}
  \caption{Complete evaluation of identity preservation in terms of face verification rates on CACD2000 \cite{chen2014cross}.}
  \label{table:idt_cacd_full}
\vspace{1cm}
  \begin{tabular}{cccc}
    \toprule
    & Average of All Pairs & Hardest Pair & Easiest Pair \\
    \midrule
    Lifespan \cite{orel2020lifespan} & 87.11\% & (test, 50-69): 72.32\% & (30-39, 50-69): 98.85\% \\
    Ours & \boldmath{$99.98\%$} & (test, 60+): 99.96\% & (test, 30-39): 100.00\% \\
    \bottomrule
     
  \end{tabular}
  \caption{Complete evaluation of identity preservation in terms of face verification rates on FFHQ \cite{karras2019style}.}
  \label{table:idt_ffhq_full}
   
\end{table*}

\end{document}